\def\year{2019}\relax
\documentclass[letterpaper]{article} 
\usepackage{aaai19}  
\usepackage{times}  
\usepackage{helvet}  
\usepackage{courier}  
\usepackage{url}  
\usepackage{graphicx}  
\usepackage{microtype}
\usepackage{csquotes}
\usepackage{ogonek}

\usepackage[export]{adjustbox}
\usepackage{amsmath, amssymb, amsthm}
\usepackage{booktabs}
\usepackage{caption}
\usepackage{color}
\usepackage{enumitem}
\usepackage{multirow}
\usepackage{subcaption}

\theoremstyle{definition}
\newtheorem{defn}{Definition}
\newcommand{\namecite}[1]{\citeauthor{#1}~\shortcite{#1}}

\begin{document}
%
\title{Measuring and Characterizing Generalization \\ in Deep Reinforcement Learning}
\author{Sam Witty\textsuperscript{1}, Jun Ki Lee\textsuperscript{2}, Emma Tosch\textsuperscript{1}, Akanksha Atrey\textsuperscript{1}, Michael Littman\textsuperscript{2}, David Jensen\textsuperscript{1} \\
  \textsuperscript{1}University of Massachusetts Amherst, 
  \textsuperscript{2}Brown University \\
}

\nocopyright

\newcommand{\gridworld}{\textsc{GridWorld}}
\newcommand{\amidar}{\textsc{Amidar}}
\newcommand{\simulator}{\textsc{Intervenidar}}
\newcommand{\TAR}{\textrm{TAR}}
\newcommand{\VEE}{\textrm{VEE}}
\newcommand{\qvalue}{Q-value}
\newcommand{\qnetwork}{Q-network}

\newcommand{\B}[1]{\fontseries{b}\selectfont#1}

\maketitle
\begin{abstract}

Deep reinforcement-learning methods have achieved remarkable performance on challenging control tasks. Observations of the resulting behavior give the impression that the agent has constructed a generalized representation that supports insightful action decisions. We re-examine what is meant by generalization in RL, and propose several definitions based on an agent's performance in on-policy, off-policy, and unreachable states. We propose a set of practical methods for evaluating agents with these definitions of generalization. We demonstrate these techniques on a common benchmark task for deep RL, and we show that the learned networks make poor decisions for states that differ only slightly from on-policy states, even though those states are not selected adversarially. Taken together, these results call into question the extent to which deep Q-networks learn generalized representations, and suggest that more experimentation and analysis is necessary before claims of representation learning can be supported.

\end{abstract}

\section{Introduction}

Deep reinforcement learning (RL) has produced agents that can perform complex tasks using only pixel-level visual input data. Given the apparent competence of some of these agents, it is tempting to see them as possessing a deep understanding of their environments. Unfortunately, this intuition can be shown to be very wrong in some circumstances.

Consider a deep RL agent responsible for controlling a self-driving car. Suppose the agent is trained on typical road surfaces but one day it needs to travel on a newly paved roadway. If the agent operates the vehicle erratically in this scenario, we would conclude that this agent has not formed a sufficiently general policy for driving. 

We provide a conceptual framework for thinking about generalization in RL. We contend that traditional notions that separate a training and testing set are misleading in RL because of the close relationship between the experience gathered during training and evaluations of the learned policy. 


With this context in mind, we address the question: 

\begin{quote}\small
To what extent do the accomplishments of deep RL agents demonstrate generalization, and how can we recognize such a capability when presented with only a black-box controller?
\end{quote}

We propose a view of generalization in RL based on an agent's performance in states it \emph{couldn't} have encountered during training, yet that only differ from on-policy states in minor ways. Our approach only requires knowledge of the training environment, and doesn't require access to the actual training episodes. The intuition is simple: To understand how an agent will perform across parts of the state space it could easily encounter and should be able to handle, expose it to states it could never have observed and measure its performance. Agents that perform well under this notion of generalization could be rightfully viewed as having mastered their environment. In this work, we make the following contributions:

    
    \textbf{Recasting generalization.} We define a range of types of generalization for value-based RL agents, based on an agent's performance in on-policy, off-policy, and unreachable states. We do so by establishing a correspondence between the well-understood notions of interpolation and extrapolation in prediction tasks with off-policy and unreachable states in RL.
    
    \textbf{Empirical methodology.} We propose a set of practical methods to: (1) produce off-policy evaluation states; and (2) use parameterized simulators and controlled experiments to produce unreachable states.
    
    \textbf{Analysis case-study.} We demonstrate these techniques on a custom implementation of a common benchmark task for deep RL, the Atari 2600 game of \amidar{}. Our version, \simulator{}, is fully parameterized, allowing us to manipulate the game's latent state, thus enabling an unprecedented set of experiments on a state-of-the-art deep \qnetwork{} architecture. We provide evidence that DQNs trained on pixel-level input can fail to generalize in the presence of non-adversarial, semantically meaningful, and plausible changes in an environment. 

\paragraph{Example.} 

\setlength{\tabcolsep}{2pt}
\begin{figure*}[!t]
\centering
\begin{tabular}[t]{@{}*{5}{c}}
    {\adjincludegraphics[width=0.35\columnwidth]{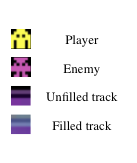}}  &
    {\adjincludegraphics[trim={11 53 11 33},clip, width=0.35\columnwidth]{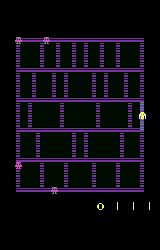}}  &
    {\adjincludegraphics[trim={11 53 11 33},clip, width=0.35\columnwidth]{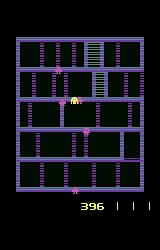}} &
    {\adjincludegraphics[trim={11 53 11 33},clip, width=0.35\columnwidth]{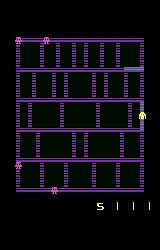}} &
    {\adjincludegraphics[trim={11 53 11 33},clip, width=0.35\columnwidth]{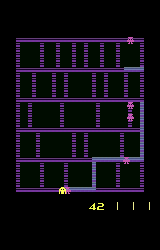}} \\
    {} & (a) Default start & (b) Default death & (c) Modified start & (d) Modified death
\end{tabular}
\caption{\label{fig:robust-behavior} A minor
change in \amidar{} game state 
can dramatically reduce a trained agent's ability to obtain a large reward. 
}
\end{figure*}

In \amidar{}, a Pac-Man-like video game, an agent moves a player around a two-dimensional grid, accumulating reward for each vertical and horizontal line segment the first time that the player traverses them. An episode terminates when the player makes contact with one of the five enemies that also move along the grid.

Consider the two executions of an agent's learned policy in Figure~\ref{fig:robust-behavior} starting from two distinct states, \emph{default} and \emph{modified}. The default condition places the trained agent in the deterministic start position it experienced during training. The modified condition is identical, except that a single line segment has been filled in. While this exact state could never be observed during training, we would expect an agent that has learned appropriate representations and a generalized policy to perform well. Indeed, with a segment filled in, the agent is at least as close to completing the level as in the default condition. However, this small modification causes the agent to obtain an order of magnitude smaller reward. Importantly, this perturbation differs from an adversarial attack~\cite{huang2017adversarial} for deep agents in that it influences the latent \emph{semantics} of state, not solely the agent's \emph{perception} of that state. Our experiments expand on this representative example, enumerating a set of agents and perturbations.

\section{Background and Related Work}

We consider the standard RL formulation of an agent sequentially interacting with an environment taking actions at discrete time steps. Formally, this process is modeled as a  4-tuple $\langle S, A, T, R\rangle$ Markov decision process (MDP). The agent starts from a state $s_0 \in S_0$ from a set of possible start states $S_0 \subset S$ and takes an action $a \in A$ at each timestep $t$. The transition function $T(s,a,s')$ is the probability of encountering state $s'$ after taking action $a$ in state $s$. The reward function $R: s \rightarrow r \in \mathcal{R}$ defines the reward the agent receives when it encounters state $s$. The agent's objective is to maximize the accumulated sum of rewards.

A policy, $\pi: s \rightarrow a$, is a mapping from states to actions, fully characterizing the behavior of an agent. The \qvalue{} of a state--action pair, $q_{\pi}(s,a)$, is the expected return for following $\pi$ from $s$ after taking action $a$, $\mathbb{E}_\pi \left[ \sum_{k=1}^{\infty} \gamma^k R(s_{t+k}) \mid s_t = s, a_t=a \right]$, where $\gamma$ is the discount rate. The value of a state, $v_{\pi}(s)$, is the expected return by following $\pi$ from $s$, $q_{\pi}(s,\pi(s))$. The optimal policy $\pi^*$ is the policy $\pi$ that maximizes $v_{\pi}(s), \forall s \in S$, which is equivalent to maximizing $q_{\pi}(s,a) \forall s, a \in S,A$.

A widely used class of methods for specifying policies in RL is to construct an approximation of the state--value function, $\hat{q}(s, a)$, and then select the action that maximizes $\hat{q}(s, a)$ at each timestep~\cite{sutton1998introduction}. Deep \qnetwork{}s (DQNs) are one such method, using multi-layer artifical neural networks as a function approximation for $\hat{q}(s, a)$. We omit discussion of recent advances in network architecture and training for brevity, as they are tangential to the core contributions of our work.

\textbf{Prior Work on Generalization in RL.} Generalization has long been a concern in RL \cite{sutton1998introduction}. Somewhat more recently, \namecite{kakade03} provided a theoretical framework for bounding the amount of training data needed for a discrete state and action RL agent to achieve near optimal reward. \namecite{nouri2009novel} discuss how to apply the idea of a training/testing split from supervised learning in the context of offline policy evaluation with batch data in RL.

Generalization has been cast as avoiding overfitting to a particular training environment, implying that sampling from diverse environments is necessary for generalization~\cite{whiteson2011protecting,zhang2018study}. Other work has focused on generalization as improved performance in off-policy states, a framework much closer to standard approaches in supervised learning. Techniques such as adding stochasticity to the policy~\cite{hausknecht2015impact}, having the agent take random steps, no-ops, steps from human play~\cite{nair2015gorila}, or probabilistically repeating the agent's previous action~\cite{machado2017revisiting}, all force the agent to transition to off-policy states. 


These existing methods diversify the training data via exposure to on-policy and off-policy states, but none discuss generalization over states that are logically plausible but unreachable. The prior focus has been on generalization as a method for preventing overfitting, rather than as a capability of a trained agent.

\textbf{Generalization vs.\ Memorization.} Generalization is often contrasted with memorization and there have been recent efforts to understand their respective roles in deep learning. For instance, with an operationalized view of memorization as the behavior of deep networks trained on noise, \namecite{arpit2017closer} showed that the same architectures that memorize noise can learn generalized behaviour on real data. 

\textbf{Adversarial Attacks on Deep Networks.} While related to adversarial attacks on deep networks, this work differs in two important ways: (1) interventions are not adversarially selected and, (2) interventions operate on latent states, not on the agent's perception. \namecite{mandlekar2017adversarially}
attempted to make agents robust to random high-level perturbations on the input. That is, for the domain they explore, MuJoCo physics simulator, the inputs are at the resolution of human-understandable concepts. Yet, this work does not address questions of alignment between meaningful real world high-level perturbations and learned representations by the network.

\section{Recasting Generalization \label{sec:generalization_RL}}

Using existing notions of generalization, such as held-out set performance, is complicated when applied to RL for two reasons: (1) training data is dependent on the agent's policy; and (2) the vastness of the state space in real-world applications means it is likely for novel states to be encountered at deployment time.

One could imagine a procedure in RL that directly mimics evaluation on held-out samples by omitting some subset of training data from any learning steps. However, this methodology only evaluates the ability of a model to \emph{use} data after it is collected, and ignores the effect of exploration on generalization. Using this definition, we could incorrectly claim that an agent has learned a general policy, even if this policy performs well on a very small subset of states. Instead, we focus on a definition that encapsulates the trained agent as a standalone entity, agnostic to the specific data it encountered during training.

\paragraph{Generalization via State-Space Partitioning.}
We partition the universe of possible input states to a trained agent into three sets, according to how the agent can encounter them following its learned policy $\pi$ from $s_0 \in S_0$. Here, $\Pi$ is the set of all policy functions, and $\alpha$, $\delta$, and $\beta$ are some small positive values close to 0. We can think of $\delta$ and $\beta$ as thresholds on estimation accuracy and optimality performance. The set of reachable states, $S_\text{reachable}$, is the set of states that an agent encounters with probability greater than $\alpha$ by following any $\pi'\in\Pi$.\footnote{These definitions can be customized with alternative metrics for value estimation and optimality, such as replacing $|\hat{v}(s) - v_{\pi}(s)|$ with $(\hat{v}(s) - v_{\pi}(s))^2$.}

\begin{defn}[Repetition]
    An RL agent has high repetition performance, $G_{R}$, if $\delta > |\hat{v}(s) - v_{\pi}(s)|$ and  $\beta > v^*(s) - v_{\pi}(s)$, $\forall s \in S_\text{on}$. The set of on-policy states, $S_\text{on}$, is the set of states that the agent encounters with probability greater than $\alpha$ by following $\pi$ from $s_0 \in S_0$.
\end{defn}

\begin{defn}[Interpolation]
    An RL agent has high interpolation performance, $G_I$, if $\delta > |\hat{q}(s,a) - q_{\pi}(s,a)|$ and  $\beta > q^*(s,a) - q_{\pi}(s,a)$, $\forall s \in S_\text{off}, a \in A$. The set of off-policy states, $S_\text{off}$, is defined as $S_\text{reachable} \setminus S_\text{on}$. 
\end{defn}

\begin{defn}[Extrapolation]
    An RL agent has high extrapolation performance, $G_E$, if $\delta > |\hat{q}(s,a) - q_{\pi}(s,a)|$ and  $\beta > q^*(s,a) - q_{\pi}(s,a)$, $\forall s \in S_\text{unreachable}, a \in A$. The set of unreachable states, $S_\text{unreachable}$, is defined as $S \setminus S_\text{reachable}$. 
\end{defn}

Note that $S$ only includes states that are in the domain of $T(s,a,s')$. In other words, specification of the transition function implicitly defines $S$, and by extension $S_{\text{unreachable}}$. This definition is particularly important in the context of deep RL, as the dimensionality of the observable input space is typically much larger than $|S|$. If we wish to demonstrate that an agent generalizes well for \amidar{}, $T(s,a,s')$ would need to be well defined with respect to latent state variables in the \amidar{} game, such as player and enemy position. If we wish to demonstrate that an agent generalizes well for all Atari games, we would need $T(s,a,s')$ to be well defined with respect to latent state variables in other Atari games as well, such as the paddle position in Breakout. Given any reaonable bound on the MDP, we would not expect the agent to perform well when exposed to random configurations of pixels.\footnote{Modifications to the transition function itself are better described as transfer learning \cite{oquab2014learning}.} 

Note that a large body of work implicitly uses $G_R$ as a criteria for performance, even though this is the weakest of generalization capabilities. It is what you get when testing a learned policy in the environment in which it was trained. Some readers may doubt that it is possible to learn policies that extrapolate well. However, \namecite{kansky2017schema} show that, with an appropriate representation, reinforcement learning can produce policies that extrapolate well under similar conditions to what we describe in this paper. What has not been shown to date is that deep RL agents can learn policies that generalize well from pixel-level input.

\begin{figure}[!t]
    \centering
    \includegraphics[width=0.8\columnwidth]{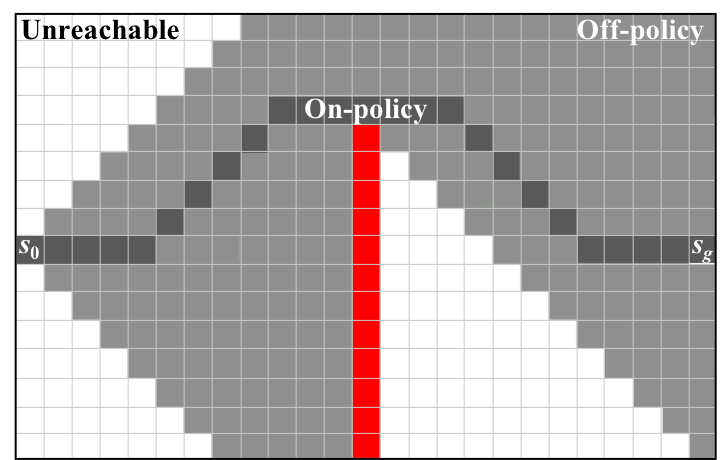}
    \caption{\label{fig:grid-world} Examples of on-policy, off-policy, and unreachable states in \gridworld{}.}
\end{figure}

We demonstrate a simple example of this state-space partition in Figure~\ref{fig:grid-world}, a classic  \gridworld{} benchmark. In this environment, the agent begins each episode in a deterministic start position, can take actions \textit{right}, \textit{right and up}, and \textit{right and down}, and obtains a reward of $+1$ when it arrives at the goal state, $s_g$. Note that the agent must move right at every step, therefore there are three regions that are unreachable from the agent's fixed start position: the upper left corner, the lower left corner, and the lower left corner after the wall. While unreachable, the upper left corner is a valid state that does not restrict the agent's ability to reach the goal state and obtain a large reward.

Note that an agent interacting in the \gridworld{} environment learns tabular \qvalue{}s, therefore we should not expect it to satisfy any reasonable definition of generalization. However, given an adequate exploration strategy, an agent could conceivably visit every off-policy state during training, resulting in $\hat{v}(s)$ converging to $v^*(s), \forall s \in S_{\text{reachable}}$. This agent would satisfy $G_R$ and $G_I$ for arbitrarily small values of $\delta$ and $\beta$. Despite this positive outcome, most observers would not say that this agent \enquote{generalizes}, because it lacks any function-approximation method. Only the definition $G_E$ is consistent with this conclusion. 

With the emergence of RL-as-a-service\footnote{e.g., https://portal.ds.microsoft.com} and concerns over propriety RL technology, evaluators may not have access to an agent's training episodes, even if they have access to the training environments. In this context, the distinction between $G_I$ and $G_E$ is particularly important when measuring an agent's generalization performance, as off-policy states may have unknowingly been visited during training. 

\paragraph{Quantifying Generalization Error.} 
Generalization in \qvalue{}-based RL can be encapsulated by two measurements for off-policy and unreachable states, one that accounts for the condition $\delta > |\hat{q}(s,a) - q_{\pi}(s,a)|$---whether the agent's estimate is close to the actual \qvalue{} after executing $\pi$---and another for the condition $\gamma > q^*(s,a) - q_{\pi}(s,a)$---whether the actual \qvalue{} is close to the optimal \qvalue{}. In our work, we use value estimate error, $\VEE_{\pi}(s) = \hat{v}(s) - v_{\pi}(s)$, and total accumulated reward, $\TAR_\pi(s) = \mathbb{E}_\pi \left[ \sum_{k=1}^{\infty} R(s_{t+k}) \mid s_t = s, a_t=a \right]$, respectively.

In most situations, $q^*(s,a)$ is not known explicitly; however, $\TAR_{\pi}(s)$ can be used to evaluate the \textit{relative} generalization ability between two agents, as the optimal value for a given state is fixed by definition. 

Unlike $\TAR_{\pi}(s)$, which, when measured in isolation can depend on the inherent difficulty of $s$, $\VEE_{\pi}(s)$ has the advantage of consistency. For example, if an agent is placed in a state such that $v^*(s) = 0$, $\TAR_\pi(s)$ alone does not capture the model's ability to generalize. $\VEE_{\pi}(s)$ may, however, if $\hat{v}(s) \approx 0$. We address this limitation of $\TAR_\pi(s)$ in our experiment by training benchmark (BM) agents on each of the evaluation conditions.

\section{Empirical Methodology}

In this section, we describe specific techniques for producing off-policy states and a general methodology for producing unreachable states based on parameterized simulators and controlled experiments.

\subsection{Off-Policy States}

It is helpful to think of off-policy states as the set of states that a particular agent \emph{could} encounter, but \emph{doesn't} when executing its policy from $s_0$. Framed in this way, the task of generating off-policy states in practice is equivalent to finding agents with policies that differ from the policy of the agent under inspection. We present three distinct categories of alternative policies for producing off-policy states, which we believe to encapsulate a broad set of historical methods for measuring generalization in RL.\footnote{We encourage readers to think critically about whether their strategy for generating off-policy states does in fact differ from the agent's policy, as this deviation may be difficult to measure.}

\textbf{Stochasticity.} One method for producing off-policy states is to introduce stochasticity into the policy of the agent under inspection~\cite{machado2017revisiting}. We present a representative method we call \emph{k off-policy actions} (k-OPA), which causes the agent to execute some sequence of on-policy actions and then take $k$ random actions to place the agent in an off-policy state. This method is scalable to large and complex environments, but careful consideration must be made to avoid overlap between states, as well as to ensure that the episode does not terminate before $k$ actions are completed. It is easy to imagine other variations, where the $k$ actions are not selected randomly but according to some other mechanism inconsistent with greedy-action selection.

\textbf{Human Agents.} The use of human agents has become a standard method in evaluating the generalization capabilities of RL agents. The most common method is known as human starts (HS) and is defined as exposing the agent to a state recorded by a human user interacting with an interface to the MDP environment~\cite{mnih2015nature}. One could easily imagine desirable variations on human starts within this general category, such as passing control back and forth between an agent and a human user. Human agents differ from other alternative agents in that they may not be motivated by the explicit reward function specified in the MDP, instead focusing on novelty or entertainment.

\textbf{Synthetic Agents.} Synthetic agents are commonly used during training in multiagent scenarios,
although to our knowledge have not been used previously to evaluate an agent's generalization ability. We present a representative method we call \emph{agent swaps} (AS), where the agent is exposed to a state midway through an alternative agent's trajectory. This method has the potential to be significantly more scalable than human starts in large and complex environments, but attention must be paid to avoiding overlap between the alternative agents and the agent under inspection. This method may also be useful in applications not amenable to a user interface or otherwise challenging to gather human data.

\subsection{Unreachable States}

Unreachable states are unlike off-policy states, which can be produced using carefully selected alternative agents. By definition, unreachable states require some modification to the training environment. We propose a methodology that is particularly well suited for applications of deep RL, where agents often only have access to low-level \emph{observable effects}, rather than what we would typically describe as a semantically meaningful or high-level representation. In the case of \amidar{} and other Atari games, for example, the position of individual entities can be described as latent state and the rendered pixels are their observable effects.

\textbf{Intervening on Latent State.} We present two distinct classes of interventions on latent state: existential, adding or removing entities, and parameterized, varying the value of an input parameter for an entity. The particular design of intervention categories and magnitude should be based on expected sources of variation in the deployment environment, and will likely need to be customized for individual benchmarks.

To facilitate this kind of intervention on latent state, we implemented \simulator{}, an \amidar{} simulator. \simulator{} closely mimics the Atari 2600 \amidar{}'s behavior,\footnote{Readers familiar with \amidar{} will know that there are other features of gameplay not listed here; although \simulator{} reproduces them, they are not important to the training regimens, nor the overall results of this paper.} while allowing users to modify board configurations, sprite positions, enemy movement behavior, and other features of gameplay without modifying \simulator{} source code. Some  manipulable features that we use in our experiments are:

\textit{Enemy existence and movement.} The five enemies in \amidar{} move at a constant speed along a fixed track. By default, \simulator{} also has five enemies whose  movement behavior is a time-based lookup table that mimics enemy position and speed in \amidar{}. Other distinct enemy movement behaviors include following the perimeter and the alternative movement protocols. These enemy behaviors are implemented as functions of the enemy's local board configuration and are used for our transfer learning experiments. 

\textit{Line segment existence and predicates.} A line segment is any piece of track that intersects with another piece of track at both endpoints. Line segments may be filled or unfilled; the player's objective is to fill all of them. In \simulator{}, users may specify which of the 88 line segments are filled at any timestep. Furthermore, \simulator{} allows users to customize the quantity and position of line segments.

\textit{Player/enemy positions.} Player and enemy entities always begin a game in the same start positions during \amidar{}, but they may be moved to arbitrary locations at any point in \simulator{}.

We included these features in the experiments because they encapsulate what we believe to be the fundamental components of \amidar{} gameplay, avoiding death and navigating the board to accumulate reward. The scale of these interventions were selected to reflect a small change from the original environment, and are detailed in the case-study section.

\textbf{Control.}
In addition to producing unreachable states, parameterizable simulators enable fine control of experiments, informing researchers and practitioners about \emph{where} agents fail to generalize, not simply that they fail macroscopically. One limitation of using exclusively off-policy states is that multiple components of latent state may be confounded, making it challenging to disentagle the \emph{causes} of brittleness from other differences between on-policy and off-policy states. Controlled experiments avoid this problem of confounding by modifying only a single component of latent state.

\begin{figure*}
    \centering
    \includegraphics{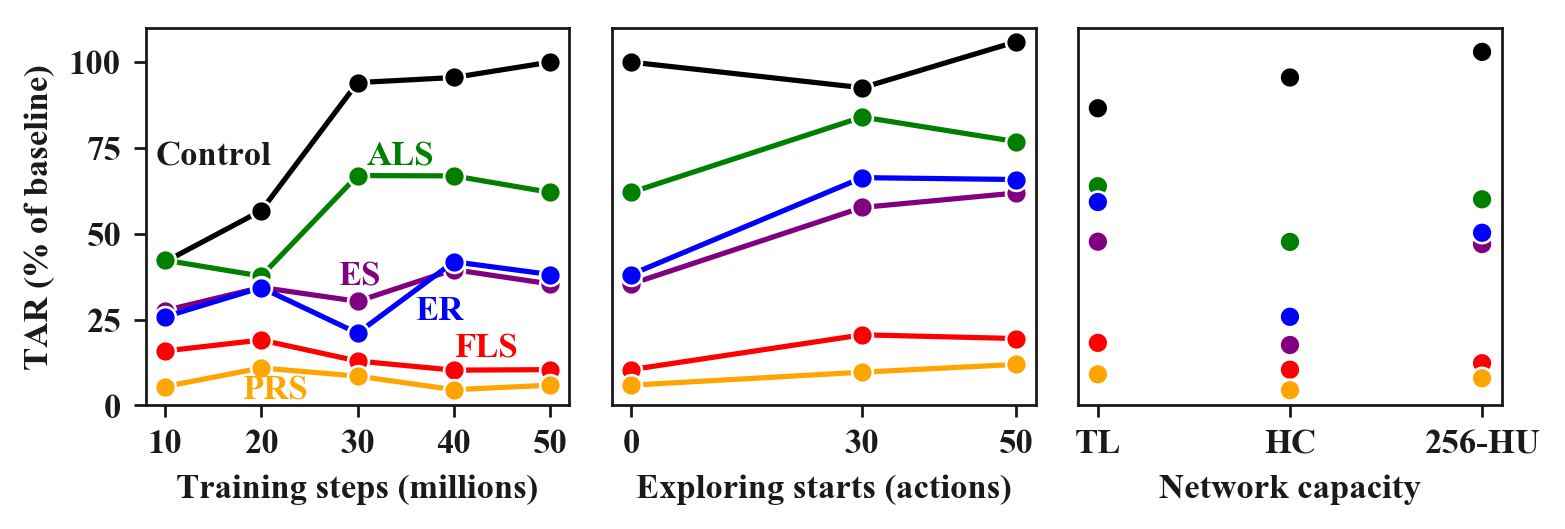}
    \caption{\label{fig:alt_agents} Average total accumulated reward (\TAR{}) from various unreachable states for each of the trained agents. The benchmark agents trained using ALS, ES, ER, FLS, and PRS configurations respectively achieved average \TAR{}s of 94, 74, 14, 77, and 90 percent of the baseline \TAR.}
    \label{fig:my_label}
\end{figure*}

    

\section{Analysis Case Study: \amidar{}}

We trained a suite of agents and evaluated them on a series of on-policy, off-policy, and unreachable \simulator{} states. Using our proposed partitioning of states and empirical methodology, we ran a series of experiments on these agents' ability to generalize. In this section, we discuss how we generated off-policy and unreachable states for the \amidar{} problem domain.

We used the standard \amidar{} MDP specification for state: a three-dimensional tensor composed of greyscale pixel values for the current, and three previous, frames during gameplay~\cite{mnih2015nature}. There are five movement actions. The transition function is deterministic, and entirely encapsulated by the \amidar{} game. The reward function is the difference between succesive scores, and is truncated such that positive differences in score result in a reward of 1. There are no negative rewards, and state transitions with no change in score result in a reward of 0.

We trained all agents using the state-of-the-art dueling network architecture, double Q-loss function, and prioritized experience replay~\cite{van2016double,wang2015dueling,schaul2016prioritized}. All of the training sessions in this paper used the same hyperparameters as in \citeauthor{mnih2015nature}'s work and we use the OpenAI's baselines implementation~\cite{baselines}.

\textbf{\amidar{} Agents.} We explored three types of modifications on network architecture and training regimens in an attempt to produce more generalized agents: (1) increasing dataset size by increasing training time; (2) broadening the support of the training data by increasing exploration at the start of each episode; and (3) reducing model capacity by decreasing network size and number of layers. To establish performance benchmarks for unreachable states, we trained an agent on each of the experimental extrapolation configurations.

\textit{Training Time.} To understand the effect of training-set size on generalization performance, we saved checkpoints of the parameters for the baseline DQN after 10, 20, 30, and 40 million training actions before the model's training reward converged at approximately 50 million actions. This process differs from increasing training dataset size in prediction tasks in that increasing the number of training episodes simulataneously changes the distribution of states in the agent's experience replay. 

\textit{Exploring Starts.} To increase the diversity of the agent's experience, we trained agents with 30 and 50 random actions at the beginning of each training episode before returning to the agent's standard $\epsilon$-greedy exploration strategy. 

\textit{Model Capacity.} To reduce the capacity of the \qvalue{} function, we explored three architectural variations from the state-of-the-art dueling architecture: (1) reducing the size of the fully connected layers by half (256-HU), (2) reducing the number of channels in each of the three convolutional filters by half respectively (HC), and (3) removing the last convolutional layer of the network (TL). Recent work on deep networks for computer vision suggest that deeper architectures produce more heirarchical representations, enabling a higher degree of generalization~\cite{krizhevsky2012ImageNet}. 

\textbf{Off-policy States.} We employed three strategies to generate off-policy states for an agent: human starts, agent swaps, and $k$-OPA. None of these methods require the \simulator{} system. In each case, we ran an agent nine times, for $n$ steps, where $n\in\{100, 200, \ldots, 900\}$.

    \textit{Human starts.} Four individuals played 30 \simulator{} games each. We randomly selected 75 action sequences lasting more than 1000 steps and extracted 9 states, taken at each of the $n$ time steps~\cite{nair2015gorila}. 

\textit{Agent swaps.}
We designated five of the trained agents as \emph{alternative agents}: (1) the baseline agent, (2) the agent that starts with 50 random actions, (3) the agent with half of the convolutional channels as the original architecture, (4) the agent with only two convolutional layers, and (5) the agent with 256 hidden units. We chose these agents with the belief that their policies would be sufficienctly different from each other to provide some variation in off-policy states.\footnote{When evaluating any of the alternative agents, we only used states from the remaining four to generate off-policy states.}

\textit{$k$-OPA.} Unlike the previous two cases where states came from sources external to the agent, in this case we had every agent play the game for $n$ steps before taking $k$ random actions, where $k$ was set to 10 and 20. 

\textbf{Unreachable States.} With \simulator{}, we generated unreachable states, guaranteeing that the agent begins an episode in a state it has never encountered during training. All modifications to the board happen before gameplay.

    \textit{Modifications to enemies.} We make one existential and one parameterized modification to enemies: We randomly remove between one and four enemies from the board (ER), and we shift one randomly selected enemy by $n$ steps along its path, where $n$ is drawn randomly between 1 and 20 (ES). 

    \textit{Modifications to line segments.} We make one existential and one parameterized modification to line segments: We add one new vertical line segment to a random location on the board (ALS) and we randomly fill between one and four non-adjacent unfilled line segments (FLS). 

    \textit{Modification to player start position.} We start the player in a randomly chosen unoccupied tile location that has at least one tile of buffer between the player and any enemies (PRS). 

\begin{figure}[t!]
    \includegraphics[width=\columnwidth]{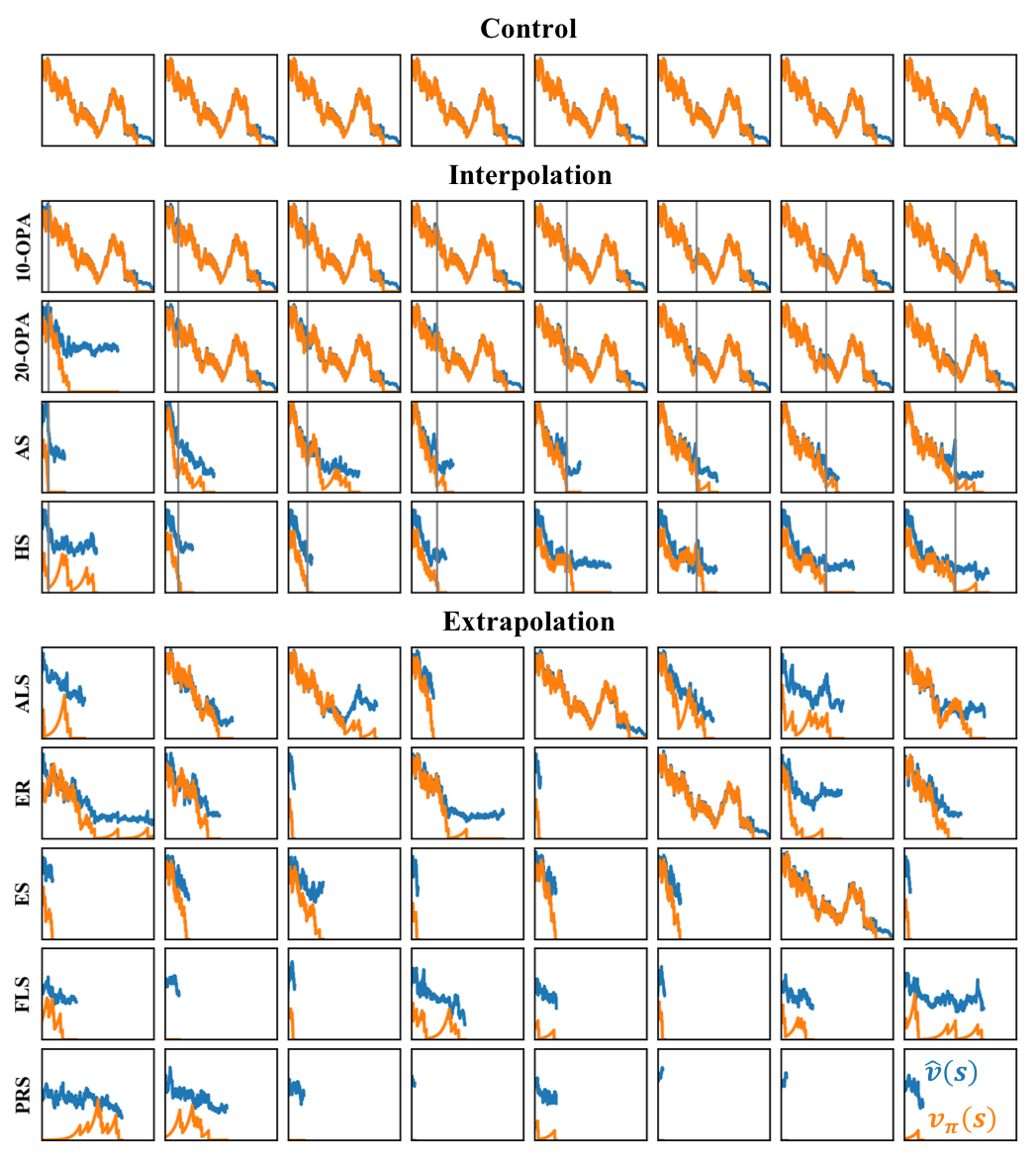}
    \caption{\label{fig:Baseline_TS} $\hat{v}(s)$ and $v_{\pi}(s)$ for replicated trajectories for all experiments. Each subplot is a single independent trial. For the interpolation experiments, the vertical grey line shows the point where the agent takes random actions (in the k-OPA experiments) or regains control (in the agent swaps and human-starts experiments). The length of each episode is consistently lower and the difference between $\hat{v}(s)$ and $v_{\pi}(s)$ is consistently higher for the extrapolation experiments.
    }
\end{figure}

\textbf{Transfer Learning: Assessing Representations.} We conducted a series of transfer learning experiments~\cite{oquab2014learning}, freezing the convolutional layers and retraining the fully connected layers for 25 million steps. We use these results to understand how learned representations in the convolutional layers relates to overall generalization performance. We train each of the agents using the alternative enemy movement protocol so that enemies move on the basis of local track features, rather than using a lookup table. If an agent has learned useful representations in the convolutional layers, then we expect that agent to learn a new policy using those representations for the alternative movement protocol.\footnote{We distinguish this transfer learning experiment from our extrapolation experiments in that the transfer learning experiment modifies the transition function $T(s,a,s')$ and by extension $q^*(s,a)$. In the extrapolation experiments, an agent can later encounter states it has observed during training and effectively use its learned policy, which is not necessarily true if the transition function changed.}

\section{Results}

Our experiments demonstrate that: (1) the state-of-the-art DQN has poor generalization performance for \amidar{} gameplay; (2) distance in the network's learned representation is strongly anti-correlated with generalization performance; (3) modifications to training volume, model capacity, and exploration have minor and sometimes counterintuitive effects on generalization performance; and (4) generalization performance does not necessarily correlate with an agent's ability to transfer representations to a new environment.

\begin{figure}
\centering
    \includegraphics[width=0.85\columnwidth]{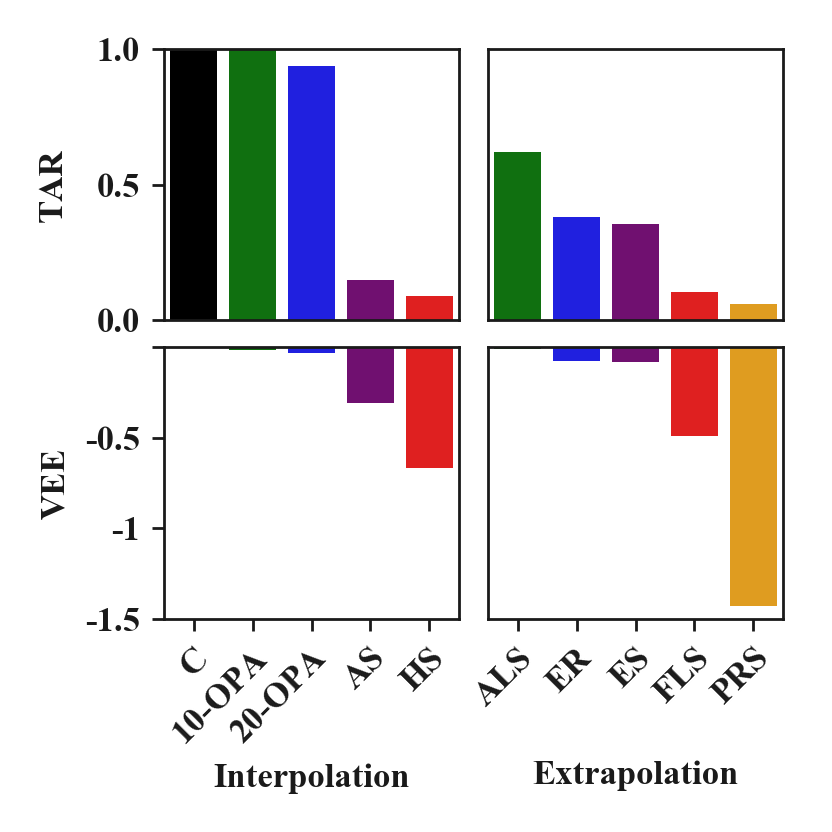}
\caption{\label{fig:Baseline_Bar} \TAR{} and average \VEE{} for control, extrapolation, and interpolation experiments. The agent consistently overestimates the state value. \TAR{} and \VEE{} are strongly anti-correlated. All \TAR{} bars are normalized by the \TAR{} of the control condition. All \VEE{} bars are normalized by their respective \TAR{}.}
\end{figure}

\textbf{Poor Generalization Performance.} Figures~\ref{fig:Baseline_TS} and~\ref{fig:Baseline_Bar} show that the fully trained state-of-the-art DQN dueling architecture produces a policy that is exceptionally brittle to small non-adversarial changes in the environment. The most egregious examples can be seen in Figure~\ref{fig:Baseline_Bar}, in the filling line segments (FLS) and player random starts (PRS) interventions. Visual inspection of the action sequences proceeeding these states showed the agent predominantly remaining stationary, often terminating the epsisode without traversing a single line segment. This behavior can be seen in Figure~\ref{fig:Baseline_TS}, where PRS and FLS episodes terminate prematurely. Videos displaying this behaviour can be found in the supplementary materials.

Furthermore, Figure~\ref{fig:Baseline_Bar} shows that \VEE{} and \TAR{} are very highly anti-correlated across the experiments, indicating that the agent's ability to select appropriate actions is related to its ability to correctly measure the value of a particular state. We observe that the model always overestimates the value of off-policy and unreachable states. In contrast,  the agent's value estimates are small and approximately symetrically distributed around 0 in the control condition.

\begin{figure}[!t]
    \centering
     \includegraphics[width=0.8\columnwidth]{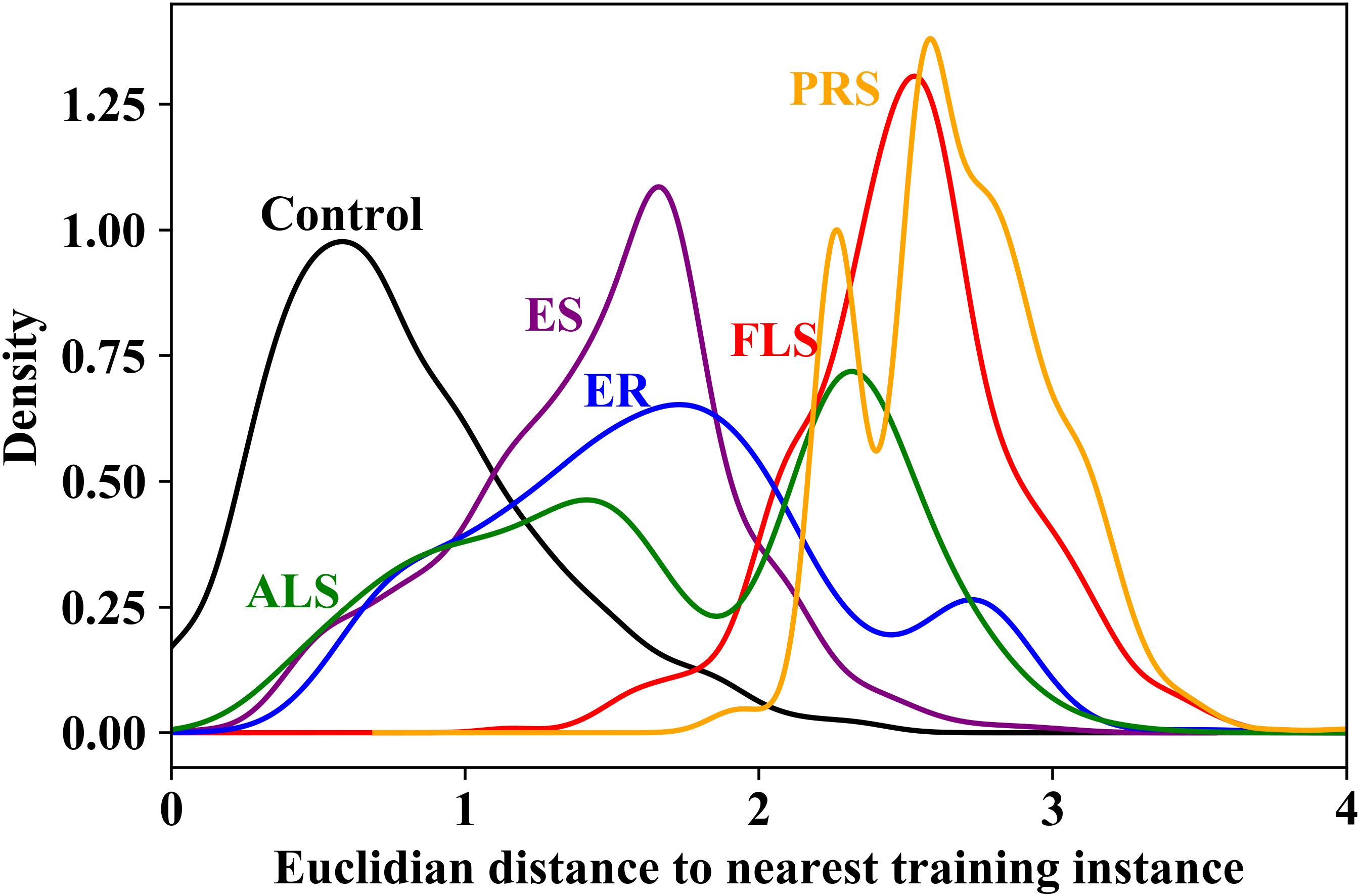}
    \caption{Smoothed empirical distributions of the distances between the test points of the extrapolation experiments and the training data. Generalization performance is anti-correlated with distance from previously seen states.}
    \label{fig:distance}
\end{figure}

\textbf{Distance in Representation.} By extracting the activations of the last layer of the DQN, we are able to observe the distance between training and evaluation states with respect to the network's learned representation. Figure~\ref{fig:distance} depicts the density estimates for the distribution of these distances. We find that the agent does not \enquote{recognize} the unreachable states where generalization is the worst, such as PRS and FLS, implying that the learned representation is inconsistent with these components of latent state. Alternatively, one could imagine a network that performs poorly by conflating states that are meaningfully different.

\textbf{Training Agents for Generalization.} We take inspiration from well-established methods in supervised learning; increasing training set size, broadening the support of the training distribution, and reducing model capacity. We propose the following analogs to each of these methods, respectively; increasing the number of training episodes, introducing additional exploration, and removing layers and nodes.

These experiments indicate that: (1) na\"{\i}vely increasing the number of training episodes until training set performance converges reduces generalization; (2) some reductions to model capacity induce improvements to generalization; and (3) increasing exploration and otherwise diversifying training experience results in more generalized policies. These results are shown in figure \ref{fig:alt_agents}.


\textit{Training Episodes.} While increasing training time clearly increases the total accumulated reward in the control condition, shorter training times appear to contribute to increased generalization ability. This increase is minimal, but it does illustrate that na\"{\i}vely increasing training time until converge of training rewards may not be the best strategy for producing generalized agents.

\textit{Model Capacity.} Of the reductions to model capacity, we find that shrinking the size of the fully-connected layers results in the greatest increase in generalization performance across perturbations. Reducing the number of convolutional layers also results in improvements in generalization performance, particularly for the enemy perturbation experiments. 

\textit{Exploration Starts.} We find that increasing the diversity of training experience has the greatest effect on generalization performance, particularly for the agent with 50 random actions. This agent experiences almost a twofold increase in total accumulated reward for human starts and all of the extrapolation experiments. This agent outperforms the baseline agent in every condition. Of particular interest is the agent's performance on the enemy shift experiments, where the agents' total accumulated reward approaches the reward achieved by an agent trained entirely in that scenario.


\textbf{Hierarchical Representations and Generalization.} While the agents with increased exploration demonstrate a clear improvement in generalization ability over baseline, it is not consistent with their ability to accumulate large reward with the alternative enemy-movement protocol after retraining. This finding contradicts those of work on representations in computer vision, where transferability of representations directly corresponds to generalization ability.

\section{Conclusions}
Generalization in RL needs to be discussed more broadly, as a capability of an arbitrary agent. We propose framing generalization as the performance metric of the researcher's choice over a partition of on-policy, off-policy, and unreachable states. Our custom, parameterizable \amidar{} simulator is a proof of concept of the type of simulation environments that are needed for generating unreachable states and training truly general agents. 

\section{Acknowledgements}
Thanks to Kaleigh Clary, John Foley, and the anonymous AAAI reviewers for thoughtful comments and contributions. This material is based upon work supported by the United States Air Force under Contract No. FA8750-17-C-0120. Any opinions, findings and conclusions or recommendations expressed in this material are those of the author(s) and do not necessarily reflect the views of the United States Air Force.





\bibliography{ref.bib}
\bibliographystyle{aaai}
\end{document}